# IMPROVING PART-OF-SPEECH TAGGING VIA MULTI-TASK LEARNING AND CHARACTER-LEVEL WORD REPRESENTATIONS

**Anastasyev D. G.** (daniil_an@abbyy.com),
**Gusev I. O.** (ilya.gusev@phystech.edu),
**Indenbom E. M.** (eugene_i@abbyy.com)
ABBYY, Moscow Institute of Physics and Technology,
Moscow, Russia

In this paper, we explore the ways to improve POS-tagging using various types of auxiliary losses and different word representations. As a baseline, we utilized a BiLSTM tagger, which is able to achieve state-of-the-art results on the sequence labelling tasks. We developed a new method for character-level word representation using feedforward neural network. Such representation gave us better results in terms of speed and performance of the model. We also applied a novel technique of pretraining such word representations with existing word vectors. Finally, we designed a new variant of auxiliary loss for sequence labelling tasks: an additional prediction of the neighbour labels. Such loss forces a model to learn the dependencies inside a sequence of labels and accelerates the process of training. We test these methods on English and Russian languages.

# УЛУЧШЕНИЕ МОРФОЛОГИЧЕСКОГО ПАРСЕРА С ПОМОЩЬЮ ВСПОМОГАТЕЛЬНЫХ ЗАДАЧ ОБУЧЕНИЯ И ПРЕДСТАВЛЕНИЙ СЛОВ НА СИМВОЛЬНОМ УРОВНЕ

**Анастасьев Д. Г.** (daniil_an@abbyy.com),
**Гусев И. О.** (ilya.gusev@phystech.edu),
**Инденбом Е. М.** (eugene_i@abbyy.com)

ABBYY, Московский Физико-Технический Институт, Москва, Россия

## 1. Introduction

A machine learning model, which just learns by heart the train examples, is hardly adequate for most purposes. Good model should rather generalize across these examples. Regularization is one of the most useful tricks, which forces models to generalize better. However, simple regularization of the model's weights usually just reduces overfitting on the train data.

In the last few years, multi-task learning with auxiliary losses became increasingly popular as a method to improve generalization achieved by the model. In this scenario, additional objectives are used during the model training. Consequently, the model's parameters are shared between different tasks and the model learns more general representations from the train dataset.

Good features are even more important for the machine learning models. In NLP tasks meaningful word representations can become such features. In most cases, we have an access to the vast unlabelled data (mostly crawled from the Internet) and by several orders smaller amount of labelled data, specific to our task. The word2vec and similar frameworks give an opportunity to pretrain word vectors on the unlabelled data. Such pretrained vectors usually improve the performance of most NLP models.

Another way to obtain the words' vectors is to use their character-level representations. Such models are usually smaller than the models with word embeddings and they do not suffer from an inability to build vector for an out-of-vocabulary word.

In this work, we discuss the ways to improve the quality of part-of-speech tagging using the auxiliary losses and propose a new variant of character-level word representations.

## 2. Baseline Model

Part-of-speech (POS) tagging is the task of assigning each word in the given text an appropriate grammatical value. Various tasks in the field of natural language processing are using the results of POS tagging. Practically all modern POS-taggers are based on recurrent neural networks (RNN). The main reason is the ability of RNNs to handle long context dependencies. It means that in contrast to more classic models where the prediction is conditioned on the narrow context window the prediction made by the RNNs is based on the whole sentence.

In case of sequence labelling problems, even more useful is the usage of Bidirectional LSTM (BiLSTM): it outputs just a concatenation of forward and backward passes of ordinary LSTMs. BiLSTMs help to use both left and right contexts information, which is usually essential for the correct analysis.

Many works showed the superiority of BiLSTMs for sequence labelling tasks. Therefore, we considered the architecture of neural network with BiLSTM in the core as our baseline. The purpose of our work was to build the best possible set of features (words' representations) for the BiLSTM and to design better objective to improve the learning process.

## 3. Word Representations

### 3.1. Word-Level Representation

Frameworks like GloVe or word2vec can build an embedding matrix with meaningful word vectors in each row. Therefore, we can map a word to the corresponding one-hot-encoding vector in very high dimensional space and multiple the embedding matrix by this vector to obtain a low-dimensional dense representation of the word.

However, the embedding matrices are typically very big: for example, the 300-dimensional embeddings of 100 thousand words have 30 million parameters. Thus, the model's whole size becomes sometimes inappropriately large.

Another problem with the word-level representations is the fixed dictionary. We are able to deal only with the words from the embeddings' vocabulary while all other words are mapped into a single unknown word vector. Apparently, we cannot store vectors for every single word: besides some rare and novel words, there always can be found some misspelt words in most texts.

### 3.2. Character-Level Representation

In order to achieve an open vocabulary and have an ability to process misspellings, one can represent word not as a single number (or one-hot-encoding vector) but as a sequence of its letters. Then some character-level function should be applied to the sequence. This function has to map an arbitrary-length sequence to a fixed dimensional vector, which can be treated as the word's representation.

The most common way to deal with such sequences of letters is to use BiLSTM. In this case, we need only the last states from both forward and backward LSTMs (Fig. 1). Such method was proven to be useful for POS-tagging by Ling (Ling, 2015).

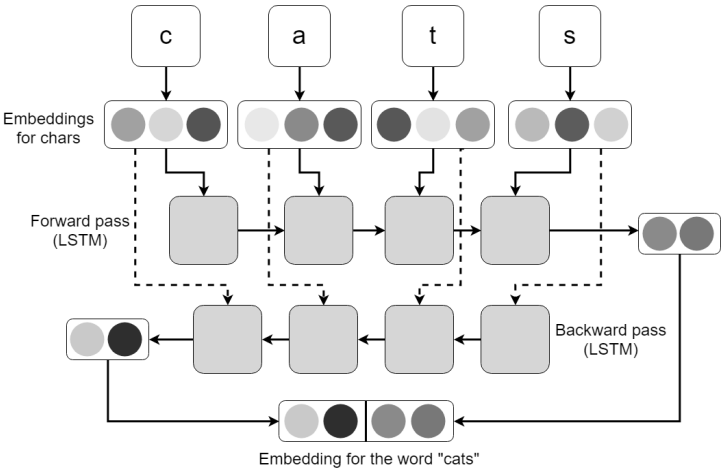

**Fig. 1.** The BiLSTM variant of the character-level word representation

We propose our own variant of leveraging the letters' sequence information. BiLSTM works much slower than a simple lookup in the embedding matrix. In order to increase the speed of computation, we are using a feedforward model (Char FF): two dense layers are applied over the concatenation of character embeddings (Fig. 3). Such representation can be computed much faster than the previous variant, though obviously slower than word embeddings. Moreover, the experiments showed that this representation can be trained much faster (in terms of the number of epochs) and with a smaller amount of the training data.

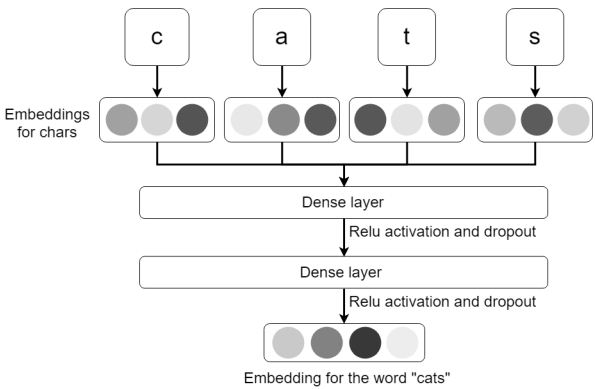

**Fig. 2.** The feedforward variant (Char FF) of the character-level word representation

The apparent disadvantage of such layer is the necessity to represent a word as a fixed length sequence. However, in our experiments, we found out that 98% of all words in the train set have a length less or equal to 11. Therefore, we added zero-padding in front of all words that are shorter than 11 symbols and cut the head of all words which are longer. Such trade-off between the speed and quality of the model and the loss of information encoded in the longer words seems reasonable to us.

### 3.3. Word Representations Pretraining

The character-level representations have much fewer parameters than the word embeddings matrices so they can be trained from scratch with the whole model. However, in this case, we are likely to suffer from overfitting. To improve the representations and utilise more unlabelled data it is possible to train such representations in the same way as the embeddings in word2vec. The disadvantage of such approach is considerably slower training process.

In order to achieve a better speed of training, we propose the following method. We are aiming to predict the word index by its representation obtained by the described variants of the character-level function. Therefore, the network consists of two parts: some character-level function, e.g. BiLSTM or the feedforward model, and the output layer with softmax activation, which predicts the word index. To ensure that the word vector predicted by such layer is meaningful we initialize the output layer by pretrained word embeddings. We used 300-dimensional GloVe vectors for this task so we added an additional dense layer, which mapped the word vector from the character-level function space to the 300-dimensional space (Fig. 3).

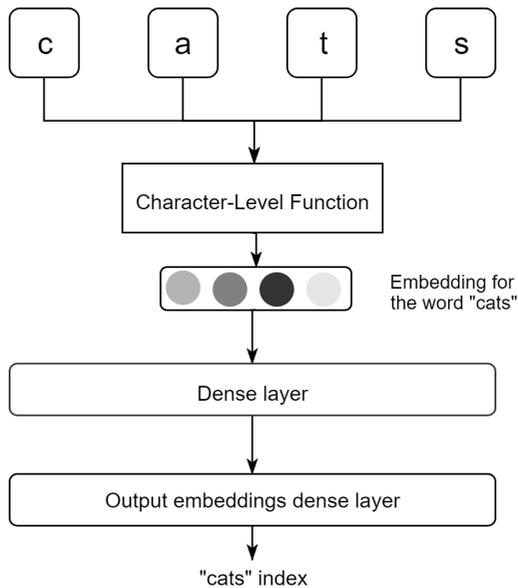

**Fig. 3.** Pretraining of the character-level functions

Mathematically the process can be seen as an optimization of the function $F$ from word characters $\bar{w} = w_1 w_2 \ldots w_n$ to 300-dimensional space:

$$\frac{\exp(w_i^T F(\bar{w}_j))}{\sum_k \exp(w_k^T F(\bar{w}_j))} \to \begin{cases} 1, i = j \\ 0, i \neq j \end{cases}$$

As a result, the representation of the word obtained by the function $F$ should be similar by cosine distance to the appropriate GloVe-vector and less similar to all other vectors.

During the designed pretraining process, we have to iterate through all words in the embeddings' vocabulary and try to minimize cross-entropy loss. To speed up this process, we used only the words from the train set. However, it is likely that optimization over all words in the pretrained word embeddings should make the obtained representation even more robust.

It is also possible to add such auxiliary loss to the model optimization process. In this case, the character-level function is optimized using two objectives: the main, task-specific objective and our auxiliary objective. The main objective forces the character-level function to produce words' representations that are suitable to the task. The auxiliary objective makes model learn representations similar to pretrained word embeddings. This should reduce model's overfitting and improve the quality of obtained representations.

Still, our experiments showed that the pretraining works better. We expect that in the case of auxiliary loss the model pays too much attention to the frequent words and ignores the infrequent ones.

### 3.4. Grammemes Embeddings

Apparently, solitary word form cannot be a good evidence for word's semantic or syntactic value: small orthographic differences may lead to completely different meanings, such as "land" vs "laud" or "taxes" vs "takes". To enhance the word's representation we propose to use grammemes embeddings.

We can represent every word with a vector where each position relates to one specific grammeme. We fill it with the estimated probability of the word to have the corresponding grammeme. For instance, the frequency of the verb "cut" is about $8.75 \cdot 10^{-5}$ its noun form has frequency $2.84 \cdot 10^{-5}$. Then we estimate the probability of the grammeme NOUN by the frequency of the noun form divided by the sum frequency of all forms of this word:

$$\frac{2.84 \cdot 10^{-5}}{2.84 \cdot 10^{-5} + 8.75 \cdot 10^{-5}} \approx 0.26$$

in this particular case.

We apply an additional linear layer with a non-linear activation on top of this vector in order to obtain not just a set of grammemes' probabilities but some interactions between them (Fig. 4). The result is similar to the feature set defined for the common linear classifiers: features as "verb + past tense" or "noun + plural" can easily be encoded by the matrix of the linear transformation. Our approach, on the other hand, gives an ability to learn some less obvious interactions between the grammemes.

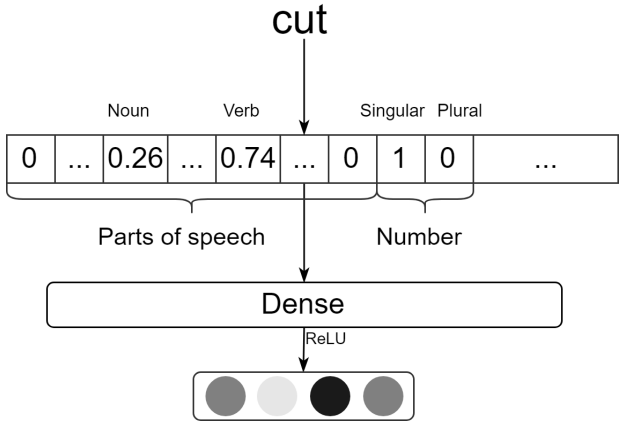

**Fig. 4.** The grammemes embedding example

We used ABBYY Compreno Morphology module to obtain the morphological analysis. It contains comprehensive dictionaries with grammatical value information and frequencies of words calculated on a large corpus.

## 4. Part-of-Speech Tagging Model Enhancements

### 4.1. Auxiliary Word Language Model

To make the training process more robust Rei (Rei, 2017) proposed to use the language modelling objective. The BiLSTM in his setup outputted hidden representations for each word, which were used not only to predict the word's tag but also its neighbours—the previous and the next words in the sentence.

However, a direct application of the sequence labelling model is prohibited because it would have an access to the full context in this case. The ordinary language model predicts the word using its left (or right) context only. To obtain similar behaviour of the sequence labelling model it's possible to use the forward LSTM only to predict the next word and backward LSTM to predict the previous one. Their hidden states then have to be concatenated to receive the whole information about both contexts, which would be used to predict the POS tag.

Such design helps to handle more general syntactic and semantic patterns in the data. Another advantage of this approach is in the clear ability to pretrain the model on a large number of unlabelled data. However, to our best knowledge, there was not any successful attempt to improve quality of any model using such pretraining process.

### 4.2. Auxiliary POS Language Model

As an alternative to the word language model, we propose to use part-of-speech language model. In our variant, we are aiming to predict the previous and the next tags using the LSTMs as well as the tag of the word.

This approach is better than the previous one in terms of training speed and required memory, because the output layer for such language model contains only tens or hundreds of elements, while in case of word language model we should consider at least ten thousand different words.

Moreover, prediction of the labels in the surrounding context forces the model to be more aware of the connections between the labels. For example, the model should learn the connection between tags "Noun" and "Adjective" or "Adverb" and "Verb".

This idea seems to be a simpler variant of structured prediction models, however, the experiments showed that its application makes a model overfit less than the model with a CRF output layer.

Nevertheless, it should be noted that it is not possible to use an unlabelled data in such variant.

### 4.3. Part-of-Speech Tagging with Transfer Learning

One of the most common ways to improve quality in the computer vision tasks is to adapt a model trained on a large dataset (usually, ImageNet) for classification task with a much smaller number of available data. Usually, the first layers of a pretrained model are frozen and only new output dense layers are trained on the task-specific data. It was shown that the first (frozen) layers are used mostly to extract useful features from the image, so they can be seen as invariant to the task.

We propose a similar idea for the POS-tagging. There are many datasets for English and Russian with different tagsets. Due to the differences in tagsets, we cannot apply a model trained on one dataset to another. Therefore, we should train a new model on the new dataset. However, we can just change the output layer and keep all other layers and their weights. Therefore, we obtain a pretrained model with meaningful weights.

To train this new model, we propose to freeze old layers during the first 5 epochs of training on the new dataset and tune only output layer. After we obtain good weights for the output layer, we can fine-tune all layers to get a better representation of the specific new corpus.

## 5. Comparison of the methods

We applied the same configuration for all the tested models and for all datasets. We used 2-layer BiLSTM with 128 units and 0.3 dropout rate. The input features were passed through the projection layer, which outputted 200-dimensional vector. As a result, all presented models had fixed size of BiLSTM layers despite the differences in the input features size. The output of the BiLSTM was projected on a lower-dimensional space using a linear layer with Batch Normalization (Ioffe, 2015) followed by the output layer with the number of units equals to the number of predicted classes.

Fig. 5 shows the final variant of the model. This variant has grammemes and character-level embeddings, CRF output layer and it was trained using POS language model auxiliary loss. The parameters of layers are specified in parentheses.

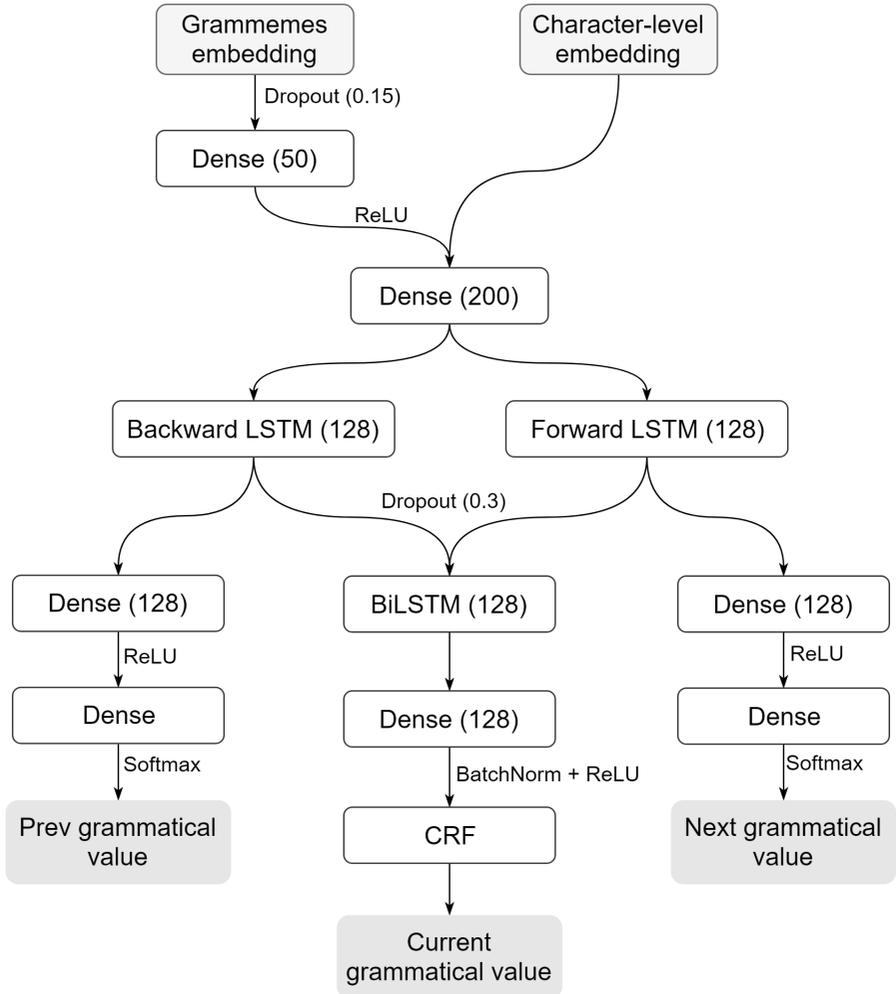

**Fig. 5.** The final model architecture.

In the next sections we are going to explore the contribution of each distinct component of the final model. We checked the quality on the Penn Treebank dataset for English language and Syntagrus (from Universal dependences 2.2) and Gikrya (from MorphoRuEval-2017) datasets for Russian. Their number of tokens are presented in Table 1.

Anastasyev D. G., Gusev I. O., Indenbom E. M.**Table 1.** Number of tokens in used corpora

| Dataset | Train | Development | Test |
|---|---|---|---|
| PTB | 912,344 | 131,768 | 129,654 |
| Syntagrus | 871,082 | 118,630 | 117,470 |
| Gikrya | 977,567 | 108,581 | 19,560 |

Penn Treebank tagset contains just 45 different tags (including punctuation tags). This corpus is somewhat standard for POS taggers evaluation—most of the well-known models were trained on it.

The Gikrya dataset has 304 grammatical values. However, during Morpho-RuEval-2017 the quality of the models was evaluated on the subset of these grammatical values (e.g., animacy category was not included into this subset). We trained our model to predict the whole set of tags but the evaluation on the test set was performed only for the specified subset (about 250 distinct grammatical values).

The Syntagrus contains 908 different grammatical values. This corpus seems better than Gikrya in terms of test set size, but we do not know any clearly state-of-the-art result shown on this corpus—which is why we evaluated our models on both corpora.

### 5.1. Word Representations

#### 5.1.1. Different Character-Level Functions

Firstly, we compared the quality of the proposed feedforward character-level function (Char FF) with more classic BiLSTM one (Char BiLSTM).

In our experiments, the Char FF model consisted of two linear layers with ReLU activations and small dropouts. The linear layers contained 500 and 200 units respectively, the dropout rate was equal to 0.15. The BiLSTM model contained 150 units. The size of character embeddings was 24.

**Table 2.** Comparison of BiLSTM and feedforward character-level models on development and test sets

| Dataset | Char BiLSTM | Char FF |
|---|---|---|
| Syntagrus | **95.23% / 95.39%** | 94.98% / 95.16% |
| Gikrya | 96.48% / **94.69%** | **96.68%** / 94.63% |
| PTB | 97.02% / 96.98% | **97.32% / 97.26%** |

These two variants of the character-level function performed roughly similar (Table 2) but the proposed feedforward model converged in a smaller number of epochs and worked much faster than BiLSTM.

#### 5.1.2. Effect of Pretraining

It seems quite reasonable to expect that the pretraining of character-level representation using the word vectors should increase the performance of the model. The experiments showed that the model with pretrained character-level function (Char FF Pretrained) has higher quality during the first few epochs and achieves accuracy about 0.1–0.2% greater than the model without pretraining (Table 3).

**Table 3.** Comparison of models with and without pretrained character-level representations

| Dataset | Char FF | Char FF (Pretrained) |
|---|---|---|
| **Syntagrus** | 94.98% / 95.16% | **95.22% / 95.36%** |
| **Gikrya** | 96.68% / **94.63%** | **96.88%** / **94.63%** |
| **PTB** | 97.32% / 97.26% | **97.40% / 97.31%** |

Such improvement leads to 4–5% error rate reduction (ERR) on Russian datasets and 2–3% ERR on PTB, which seems significant enough given the simplicity and cheapness of the pretraining process.

### 5.1.3. Grammemes Embeddings

The models with grammemes embeddings converge much faster. Grammemes embeddings seriously improved the models' performance on both Russian datasets. However, the English model without them achieved approximately the same quality after a large number of epochs (on the other hand, the model with grammemes embeddings needed roughly twice as lesser epochs to converge).

**Table 4.** Comparison of models with and without grammemes embeddings

| Dataset | Char FF (Pretrained) | + Grammemes |
|---|---|---|
| **Syntagrus** | 95.22% / 95.36% | **96.77% / 97.00%** |
| **Gikrya** | 96.88% / 94.63% | **98.07% / 95.36%** |
| **PTB** | 97.40% / **97.31%** | **97.43%** / 97.30% |

Obtained results may be explained by the differences in Russian and English morphologies. Russian language has considerably more complicated morphological system. As a result, Russian grammemes embeddings are far more informative. Moreover, it is usually harder to predict morphological tag for Russian words using merely word's form.

### 5.2. Language Model Auxiliary Objectives

Both word and POS auxiliary objectives effectively work as regularizer: a model with them overfits considerably slower and achieves better results. Slower overfitting in our case means that the difference between train and development accuracies remained insignificant even after 200 epochs, while without such auxiliary objective this difference became more than 1% after the first 100 epochs on all datasets.

The achieved results are presented in Table 5. Word and POS LM objectives gave similar results on the Penn Treebank. However, POS LM has shown clearly better performance on the Russian datasets.

Anastasyev D. G., Gusev I. O., Indenbom E. M.**Table 5.** Comparison of models with different auxiliary objectives

| Dataset | Char FF (Pretrained) + Grammemes | + Word LM | + POS LM |
|---|---|---|---|
| **Syntagrus** | 96.77% / 97.00% | 96.69% / 96.96% | **96.97% / 97.24%** |
| **Gikrya** | 98.07% / 94.85% | 97.91% / 96.30% | **98.12% / 96.72%** |
| **PTB** | 97.43% / 97.30% | 97.57% / 97.49% | 97.57% / 97.49% |

The most significant improvement was achieved on the Gikrya test set. Yet, the 7% ERR on the PTB test set and 8% ERR on Syntagrus test seem good enough to consider the proposed auxiliary objective successful.

### 5.3. CRF Layer

Usage of CRF output layer usually leads to noticeable improvement in the model's quality. However, we could not achieve better results with the CRF layer. We expect that the main reason is our POS LM objective: it forces models to learn the same connections between the tags as CRF layer does.

**Table 6.** Comparison of models with and without CRF layer

| Dataset | Char FF (Pretrained) + Grammemes + POS LM | + CRF |
|---|---|---|
| **Syntagrus** | **96.97% / 97.24%** | 96.72% / 96.97% |
| **Gikrya** | **98.12% / 96.72%** | 98.07% / 96.65% |
| **PTB** | 97.57% / 97.49% | **97.60% / 97.51%** |

### 5.4. Transfer Learning

Finally, we applied the proposed transfer learning process. We pretrained two models. The first one was trained on Compreno corpus with about 10 million tokens. The tagset contains 1040 different grammatical values. The second one was trained on the Gikrya corpus. The results are presented in Table 7.

**Table 7.** Qualities of pretrained models on Syntagrus corpus

| Model | Accuracy |
|---|---|
| **Base** | 96.97% / 97.24% |
| **Compreno pretrained** | 98.18% / 98.29% |
| **Gikrya pretrained** | **98.21% / 98.33%** |

Gikrya and Syntagrus have similar tagset (based on Universal Dependencies standard) while Compreno's tagset and applied conventions are very different from Syntagrus corpus. Therefore, it is understandable that Gikrya pretrained model achieved better quality.

As a result, we achieved very significant improvement using quite a simple process.

## 5.5. Summary

To conclude, we applied few tricks to achieve results that are on par with state-of-the-art results on considered datasets. We did not try to optimize the hyper-parameters of the models so the work should be seen mostly as a proof-of-concept.

**Table 8.** Comparison with the existing models on PTB dataset

| Tagger | Test Acc |
|---|---|
| Manning (2011) | 97,32% |
| Søgaard (2011) | 97,50% |
| Santos (2014) | 97,32% |
| Ling (2015) | **97,78%** |
| Ma (2016) | 97,55% |
| Choi (2016) | 97,64% |
| Rei (2017) | 97,43% |
| This work | 97,51% |

**Table 9.** Comparison with results on MorphoRuEval-2017 on Gikrya dataset

|  | Modern literature | News | VKontakte |
|---|---|---|---|
| **Best closed track model** | 94.16% | 93.71% | 92.29% |
| **Best open track model** | **97.45%** | 97.37% | **96.52%** |
| **Our model** | 96.46% | **97.97%** | 95.64% |

The final model is worse than the best PTB models. On the other hand, it does not use word embeddings. That means that our model is much smaller. To our best knowledge, the achieved result is the best for models without word embeddings.

The model also shows poorer performance than the best model on MorphoRuEval-2017. However, the best model was trained using additional data and it used word embeddings too.

## 6. Conclusions

We proposed a new method of word's representation using character-level feed-forward neural network and a way to pretrain such representations with existing word embeddings. This variant of words representations seems to perform better than the previous ones. Moreover, the pretraining process helps to use the information encoded in the word embeddings implicitly. That means we can expect syntactic and semantic meaningfulness of the representations, which can be found in the word2vec vectors.

We described a way to encode an additional information about the grammatical value of the word in the grammemes embedding. Such embeddings are useful as a way to regularise the word representation and provide it with an additional morphological information.

We also proposed a novel approach to multi-task learning for sequence labelling tasks and tested it on POS tagging problem. The achieved results a comparable to the state-of-the-art in this area.

Finally, we proved the possibility of transfer learning for the POS tagging task. We achieved almost 40% ERR on Syntagrus dataset by applying this process.

Summing up, the proposed tricks are aimed to improve the quality of input features and to regularize the learned objective. We consider them simple enough to be applicable for most of NLP tasks and the achieved results seem reasonably good.

## References

1. *Anastasyev D. G. et al.* (2017) Part-of-speech tagging with rich language description, Computational linguistics and intellectual technologies: Proceedings of the International Conference "Dialog 2017". Vol. 1, pp. 2–13
2. *Anisimovich K. V., et al.* (2012), Syntactic and Semantic Parser Based on ABBYY Compreno Linguistic Technologies, Computational linguistics and intellectual technologies: Proceedings of the International Conference "Dialog 2012". Vol. 2, pp. 91–103
3. *Choi J.* (2016), Dynamic Feature Induction: The Last Gist to the State-of-the-Art, Proceedings of the 54th Annual Meeting of the Association for Computational Linguistics (NAACL 2016), San Diego, CA, pp. 271–281.
4. *Hochreiter S., Schmidhuber J.* (1997), Long Short-Term Memory, Neural Computation, Vol. 9, Issue 8, pp 1735–1780.
5. *Huang Z., Xu W., Yu K.* Bidirectional LSTM-CRF models for sequence tagging, arXiv:1508.01991.
6. *Ioffe S., Szegedy Ch.* (2015), Batch Normalization: Accelerating Deep Network Training by Reducing Internal Covariate Shift, arXiv:1502.03167
7. *Korobov M.* (2015), Morphological Analyzer and Generator for Russian and Ukrainian Languages, Analysis of Images, Social Networks and Texts, Vol. 542, pp. 320–332.
8. *Ling W. et al* (2015), Finding Function in Form: Compositional Character Models for Open Vocabulary Word Representation, available at: https://arxiv.org/abs/1508.02096.
9. *Ma X., Hovy Ed.* (2016), End-to-end Sequence Labeling via Bi-directional LSTM-CNNs-CRF, arXiv:1603.01354.
10. *Rei M.* (2017), Semi-supervised Multitask Learning for Sequence Labeling, arXiv.org:1704.07156.
11. *Søgaard A.* (2011), Semisupervised condensed nearest neighbor for part-of-speech tagging, Proceedings of the 49th Annual Meeting of the Association for Computational Linguistics, pp. 48–52
12. *Sorokin A., et al.* (2017) MorphoRuEval-2017: an Evaluation Track for the Automatic Morphological Analysis Methods for Russian, Proceedings of the International Conference "Dialog 2017". Vol. 1, pp. 297–313